\documentclass{llncs}

\usepackage[utf8]{inputenc}
\usepackage[T1,T2A]{fontenc}

\usepackage[final]{graphicx}
\usepackage{epstopdf}
\usepackage[labelsep=period]{caption}
\usepackage[hyphens]{url}
\usepackage[]{algorithm2e}
\usepackage{verbatim}

 \usepackage[russian,english]{babel}

\begin{document}

\title{Random Forest Based Approach for Concept Drift Handling}

\author{
Aleksei V.  Zhukov\inst{1,2}\\
Denis N. Sidorov\inst{1,2}
\and
 Aoife M. Foley\inst{3}
}

\institute{
Energy Systems Institute SB RAS, Irkutsk, Russia\\
\email{dsidorov@isem.irk.ru},\\
\and
Irkutsk State University, Irkutsk, Russia\\
\email{zhukovalex13@gmail.com}
\and
Queens University Belfast, Belfast, UK\\
\email{a.foley@qub.ac.uk }
}

\maketitle

\begin{abstract}
Concept drift has potential in smart grid analysis because the socio-economic behaviour 
of consumers is not governed by the laws of physics. Likewise there are also applications 
in wind power forecasting.
 In this paper we present decision tree ensemble classification method based on the Random Forest algorithm for concept drift.
 The weighted majority voting ensemble aggregation rule is employed based on the ideas of Accuracy Weighted Ensemble (AWE) method. Base learner weight in our case is computed for each sample evaluation using base learners accuracy and intrinsic proximity measure of Random Forest. Our algorithm exploits both temporal weighting of samples and ensemble pruning as a forgetting strategy.
 We present results of empirical comparison of our method with оriginal random forest with incorporated ``replace-the-looser'' forgetting andother state-of-the-art concept-drfit classifiers like AWE2.

\vspace{1em}
\textbf{Keywords:} machine learning, decision tree, concept drift, ensemble learning, classification, random forest.
\end{abstract} 

\section{Introduction}

Ensemble methods of classification (or, briefly, ensembles) employ various learning algorithms to obtain better predictive accuracy comparing with individual classifiers. 
Ensembles are much used in research and most of the research is devoted to stationary environments where the complete datasets are available for learning
classifiers and transfer functions of dynamical systems are not changing  as time goes on. For  real world applications (e.g. in power engineering \cite{arxiv2016}, \cite{ijai2015})
 learning algorithms are supposed to work in dynamic environments with data continuously generated in the form of a stream on not necessarily equally spaced time intervals.
 Data stream processing commonly relies on single scans of the training data and  implies restrictions on memory and time.  Changes caused by dynamic environments (e.g. consumer behaviour in future smart grids) can be categorised into sudden or
gradual concept drift subject to appearance of novel classes in a stream and
the rate of changing definitions of classes. 

One of the most generally effective ensemble classifier is a Random Forest. This algorithm employs bagging~\cite{brieman} and principles of the random subspace method~\cite{ho} to build a highly decorrelated ensemble of decision trees~\cite{brieman2001}. There are some attempts \cite{saffari2009online} to adapt Random Forest to handling concept-drift, but this approach is not fully discovered.

The objective of this paper is to propose a novel approach to classification  to adapt to concept drifts. This novel classifcation method can be applied to consumer behaviour in future
smart grids to predict and classify the random behaviour of humans when they interact with 
smart appliances in the home and charging and discharging of electric vehicles. Likrwise in these methods can be used to forecast windpower and indeed solar power.

We propose to compute the base learner weight for samples evaluation using base learners accuracy and intrinsic proximity measure of random forest.

It is to be noted that concept drifting is related to non-stationary dynamical systems modelling~\cite{sid2002}, where  the completely supervised learning is employed using a special set of input test signals. For more details concerning integral dynamical models theory and applications refer to monograph~\cite{sid2015} and its bibliography.

This paper is organised into five sections. In  Section~\ref{sec:relatedWork}, a brief review of ensemble streaming classifiers based on random forest is given. Section~\ref{sec:proposedAlgorithm} delivers detailed presentation of proposed algorithm. In Section~\ref{sec:experiments} experiments on both machine learning and tracking tasks are provided. Finally, the paper concludes with Section~\ref{sec:conclusions} where the main results  are discussed.

\section{Related Work}
\label{sec:relatedWork}

The use of drifting concepts  for huge datasets analysis is not unfamiliar to the machine learning and systems identification communities \cite{ibm2003}. 
In this work we restrict ourselves to considering  decision tree ensamble classification methods only. 

Let us briefly discuss methods most related to our proposal and employed in the experiments.
 For a more detailed overview of the results in 
this area, including online incremental ensembles, readers may refer to  monograph \cite{monogama} and review \cite{kunchevaOverview}. Bayesian logistic regression
was used in~\cite{krasot2013} to handle drifting concepts in terms of dynamical programming.
Concept drifting handling is close to methodology of on-line random forest
algorithm \cite{saffari2009online} where ideas from on-line bagging,
extremely randomised forests and  on-line decision
tree growing procedure are employed.
 Accuracy Weighted Ensemble (AWE) approach was proposed in \cite{ibm2003}. The main idea is to train a new classifier on each incoming dataset and use it to evaluate all the existing classifiers in the ensemble.  We incorporate this idea in 
our approach.

A Massive Online Analysis (MOA) framework \cite{MOA} is the one of the popular benchmarks for testing online classification, where clusterization and regression algorithms are written in Java. Java software contains the state of the art classifiers for concept drift handling such as SEA \cite{sea} and Online Bagging \cite{obag}. 

Analysis of ensemble methods with decision tree base learners is interesting topic to be addressed in this paper. The recursive nature of decision trees
makes on-line learning a difficult task due to the hard splitting rule, errors cannot be corrected further down
the tree.

\section{Proximity Driven Streaming Random Forest}
\label{sec:proposedAlgorithm}

In paper \cite{wang2003mining} the authors clearly demonstrated that a classifier ensemble can outperform a single classifier in the
presence of concept drifts when the base classifiers of the ensemble are adherence weighted  to the error similar to current testing samples. We propose another approach which exploits   Random Forest properties.

To produce a novel algorithm capable of handling concept drift the following questions need to be answered:
\begin{itemize}
	\item How to adapt original Random Forest for data streaming?
	\item How to define sample similarity metric?
	\item How to choose the base classifier weighting function?
	\item How to choose forgetting strategy?
\end{itemize}
These questions are considered in the following subsections.

\subsection{Streaming classifier based on Random Forest}

Methodologically ensemble approaches allow  concept-drift to be handled in the following ways: base classifier adaptation, changing in training dataset (such as Bootstrap \cite{brieman} or RSM \cite{ho}), ensemble aggregation rule changing or changing in structure of an ensemble (pruning or growing).
In this paper we propose Proximity Driven Streaming Random Forest (PDSRF) which exploit combinations of these approaches. Besides some methods are already incorporated to the original Random Forest. Contrary to conventional algorithms we use weighted majority voting as an aggregation rule of ensemble. This allows us to adapt the entire classifier by changing the weights of the base learners.
In order to obtain the classifiers weight estimation we should store samples. For this purpose we use a sliding windows approach which is used in the periodicaly updated Random Forest \cite{mmro2015zhukov}. The length of this window is fixed and can be estimated by cross-validation.
Random Forest \cite{brieman2001} uses unpruned CART \cite{breiman1983cart} trees and for $N$ instances and $Mtry$ atributes chosen at each decision tree node average computational cost of ensemble building is $O(T Mtry N log^2 N)$, where $T$ is a number of trees. It can be unsifficient for online applications. To reduce the comlexity we use the randomization approach proposed in Extremely Randomized Trees \cite{geurts2006extremely}. In our implementation the split set consists of randomly generated splits and the best one is choosen by minimization of the Gini-index measure. So that $O(T Mtry N logN)$ cost complexity can be achieved.

\subsection{Sample similarity metric}

We employ the assumption that the base classifiers make similar errors on similar samples even under concept-drift.  
Conventional Random Forest exploy the so called proximity measure. It uses a tree structure to obtain similarity in the following way: if two different sample are in the same terminal node, their proximity will be increased by one. At the end, proximities are normalised by dividing by the number of trees~\cite{brieman2001}.

\subsection{Base classifier weighting function}

Following the AWE approach proposed in \cite{wang2003mining} we use an error rate to produce weights (\ref{equ:errorweight}) of classifiers, where $E$ is an new block testing error for $i$-th classifier, $\varepsilon$ is a small parameter.

\begin{equation}
 w_i = 1/( E^2 + \varepsilon )
\label{equ:errorweight}
\end{equation}

\subsection{Forgetting strategy}

One of the main problems in concept-drifting learning is to select the proper forgetting strategy and forgetting rate \cite{kunchevaOverview}. The classifier should be  adaptive enough to handle changes. In this case different strategies can be more appropriate to different types of drift (for example, sudden and gradual drifts). In this paper we focus on gradual changes only.

We propose two different ways to handle the concept-drift:
\begin{itemize}
	\item temporal sample weighting,
	\item ensemble pruning technique.
\end{itemize}

\subsubsection{Data forgetting through temporal sample weighting.}

We use sample weighing to decrease the influence of old samlpes to learn the new trees  (\ref{equ:tempweights}).

\begin{equation}
w_x(t) = exp(-\alpha t)
\label{equ:tempweights}
\end{equation}
where $\alpha$ is the sample weighting rate which can be selected in experiment by cross-validation.

\subsubsection{Knowledge forgetting through ensemble growing and pruning}

In this part we apply the classic \textit{replace-the-looser} approach~\cite{kunchevaOverview} to discard trees with high error on new block samples.

\subsection{Algorithm}

To predict the sample we  Algorithm~\ref{algo:pdsrfPredict}. First we use a stored window to find similar items using the specified similarity metric. Second we evaluate our current ensemble on similar examples. Then we compute weights adherence to errors on $k$ similar samples. 

On every chunk the algorithm tests all the trees to choose the poorest base learner and replace it with new one trained on new block data. This process is iterative while the ensemble error on new block samples is higher than a specified threshold.

\begin{algorithm}
	\DontPrintSemicolon 
	\KwIn{	\\
		$S$: data stream chunks which sequentially produce training examples $\langle x, y \rangle$ \\
		$S_i$: data stream chunks of $blockSize$  \\
		$k$: number of the nearest neighbours\\
		$W(E)$: Classifier weighting function\\
	}
	\KwOut{Class probability vector}

	$nearest \gets$ find $k$ nearest samples from cache using proximity metric.
	
	\For{ \textbf{all} $x_i \in S_i$} {
			\For{ \textbf{all} trees $c_i$ in ensemble} {
				Get average error $E^{nearest}_{i}$ of $c_i$ on nearest samples
				Add weighted by (\ref{equ:errorweight}) base classifier probability to class probability vector 	
			}
	}
	
	\Return{Probability Vector}
	\caption{{\sc PDSRF} prediction algorithm}
	\label{algo:pdsrfPredict}
\end{algorithm}

\section{Experimental Evaluation}
\label{sec:experiments}

In this experiment, we evaluate our algorithm on various publicly available datasets like CoverType~\cite{dataset},  and
compared it to the most popular concept-drift classifiers. We compare our algorithm with SEA~\cite{sea}, Hoeffding Adaptive tree~\cite{hat}, Online Bagging~\cite{obag} and AWE~\cite{ibm2003} implemented in MOA~\cite{MOA}. Our algorithm was implemented natively in C++  according to the same testing methodology. Extremely randomized trees is used as a base learner. With adherence to this methodology classication accuracy was calculated using the \textit{data block evaluation method}, which exploits the \textit{test-then-train} approach. The data block evaluation method reads incoming samples without processing them, until they form a data block of size $d$. Each new data block is first used to test the existing classifier, then it updates the classifier \cite{brze2010}.

Proposed \,proximity \,driven\, streaming \,\,random forest \,\,was\,\,\, tested \,with $blockSize = 300$, $windowSize = 1500$,
number of nearest neighbours is 20 and ensemble consists of 30 decision trees. In order to  make the results more interpreteble we also test оriginal random forest with incorporated \textit{``replace-the-looser''} forgetting.

\subsection{Datasets}

We follow the literature on adaptive ensemble classifiers and select the publicly available benchmark datasets with concept drift. In this paper the proposed method was tested on the {\it cover type dataset} from Jock A. Blackard, Colorado State University \cite{dataset}.
{\it Cover type dataset} contains the forest cover type for $30 \times 30$ meter cells obtained from US Forest Service Region 2 Resource Information System (RIS) data. It contains $581\, 012$ instances and $54$ attributes, and it has been used as a benchmark in several papers on data stream classification.

\subsection{Results}

\begin{figure}[htbp]
		\centering
		\includegraphics[scale=0.35]{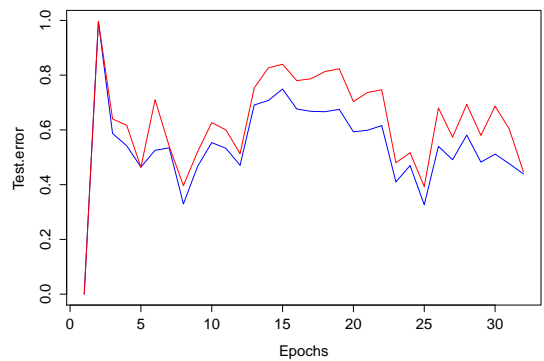}
		\caption{Original random forest with incorporated ``replace-the-looser'' (blue) and\,\,\,\,\,\, PDSRF (red) accuracy.}
		\label{fig:img_error}
	\end{figure}

The proposed approach shows results similar to the above mentioned AUE2 \cite{aue2}.
Next comparison (ref. Fig. \ref{fig:img_error}) of the original random forest with incorporated ``replace-the-looser'' forgetting and proposed proximity driven streaming random forest are presented. In Table \ref{tab:validation} the mean accuracy is shown.

\begin{table}[t]
\begin{center}
\begin{tabular}{ | l | c | } \hline
 Method & Mean accuracy\\ \hline \hline
 RF & 0.54 \\ \hline
 PDSRF & 0.63\\ \hline
 AVE2(buffered) & 0.81\\ \hline
\end{tabular}%
\end{center}
\caption{Cover type dataset mean accuracy results }
\label{tab:validation}%
\end{table}

\begin{remark}
For sake of space, we do not report the complete tests in this paper.
For more details are complete results of our approach evaluation readers may refer to  \url{http://mmwind.github.io/pdsrf}.
\end{remark}

\section{Discussion and Conclusions }
\label{sec:conclusions}
In comparison to other concept-drift approaches like Online Random Forest and the AWE, our approach needs more computational resources and  thus more time for both the training and prediction stages. But the proposed approach is highly parallelisable and can be implemented using the GPGPU.
It must be noted that the proposed approach can be efficiently applied only to  gradual concept drifts. PDSRF is senisible to all the parameters changes and all of these parameters must be accurately tuned.
As it shown the proposed approach significantly exceeds the original random forest with incorporated ``replace-the-looser'' forgetting. Although the presented results show that the accuracy is lower than AWE2, the approach has some promising directions is that the Random Forest can be used in unsupervised and allows to work  with missing data, which is an issue
with smart grid datasets and wind power forecasting where telecommunications signals 
and data recording is not 100\% robust.s

\bigskip
\subsubsection*{Acknowledgment.}
This work is partly funded by the RSF grant  No. 14-19-00054.

\bibliographystyle{splncs}
\bibliography{biblio}

\end{document}